\def\year{2020}\relax
\documentclass[letterpaper]{article} % DO NOT CHANGE THIS
\usepackage{aaai20}  % DO NOT CHANGE THIS
\usepackage{times}  % DO NOT CHANGE THIS
\usepackage{helvet} % DO NOT CHANGE THIS
\usepackage{courier}  % DO NOT CHANGE THIS
\usepackage[hyphens]{url}  % DO NOT CHANGE THIS
\usepackage{graphicx} % DO NOT CHANGE THIS
\usepackage{comment}
\usepackage{booktabs}
\usepackage{mathtools}
\usepackage{wrapfig}
\usepackage{amssymb}
\usepackage{xcolor}

\newcommand{\eg}{\textit{e}.\textit{g}.}
\newcommand{\ie}{\textit{i}.\textit{e}.}

\newcommand{\inlineimg}[1]{\raisebox{-0.2\baselineskip}{\includegraphics[height=0.95\baselineskip]{#1.png}}}
\newcommand{\inlinename}[1]{\raisebox{-0.2\baselineskip}{\includegraphics[height=0.85\baselineskip]{#1.png}}}
\newcommand{\inlinepe}[1]{\raisebox{-0.2\baselineskip}{\includegraphics[height=0.9\baselineskip]{#1.png}}}

\title{General Partial Label Learning via Dual Bipartite Graph Autoencoder}
\author{
Brian Chen,\textsuperscript{\rm 1}
Bo Wu,\textsuperscript{\rm 1}
Alireza Zareian,\textsuperscript{\rm 1}
Hanwang Zhang,\textsuperscript{\rm 2}
Shih-Fu Chang\textsuperscript{\rm 1}\\
\textsuperscript{\rm 1}Columbia University,
\textsuperscript{\rm 2}Nanyang Technological University\\
\{bc2754,bo.wu,az2407,sc250\}@columbia.edu; hanwangzhang@ntu.edu.sg
}

\begin{document}

\maketitle

\begin{abstract}
We formulate a practical yet challenging problem: General Partial Label Learning (GPLL). Compared to the traditional Partial Label Learning (PLL) problem, GPLL relaxes the supervision assumption from \emph{instance-level} --- a label set partially labels an instance --- to \emph{group-level}: 1) a label set partially labels a group of instances, where the \emph{within-group} instance-label link annotations are missing, and 2) \emph{cross-group} links are allowed --- instances in a group may be partially linked to the label set from another group. Such ambiguous group-level supervision is more practical in real-world scenarios as additional annotation on the instance-level is no longer required, \eg, face-naming in videos where the group consists of faces in a frame, labeled by a name set in the corresponding caption. In this paper, we propose a novel graph convolutional network (GCN) called Dual Bipartite Graph Autoencoder (DB-GAE) to tackle the label ambiguity challenge of GPLL. First, we exploit the cross-group correlations to represent the instance groups as \emph{dual bipartite graphs}: within-group and cross-group, which reciprocally complements each other to resolve the linking ambiguities. Second, we design a GCN autoencoder to encode and decode them, where the decodings are considered as the refined results.
It is worth noting that DB-GAE is \emph{self-supervised} and \emph{transductive}, as it only uses the group-level supervision without a separate offline training stage. Extensive experiments on two real-world datasets demonstrate that DB-GAE significantly outperforms the best baseline over absolute 0.159 F1-score and 24.8\% accuracy. We further offer analysis on various levels of label ambiguities.
\end{abstract}

\section{Introduction} \label{sec:introduction}
%Background of ambiguous data learning and PLL problem

 \begin{figure}[!th]
 \resizebox{\columnwidth}{!}{
  \includegraphics[width=\columnwidth]{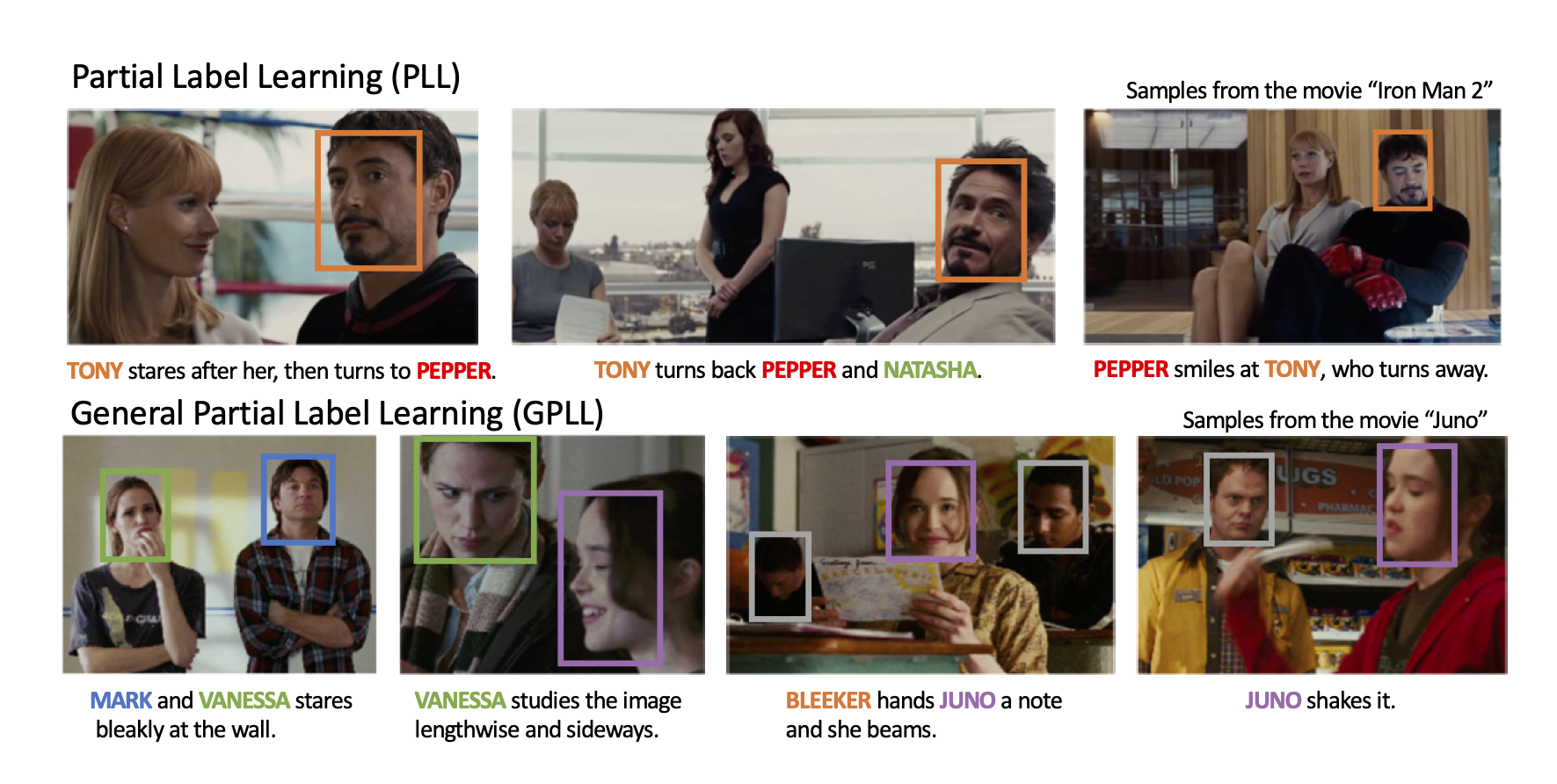}}
  %\vspace{-5mm}
\caption{\textbf{Top:} Traditional PLL problem --- each face is associated with one of the names in the caption. \textbf{Bottom:} The proposed GPLL problem addressed in this paper --- there are general cases of faces without names or names without faces.  See Figure 2 for illustrative formulation differences. Images are from MPII-MD ~\cite{RohrbachCVPR2017} and M-VAD ~\cite{pini2019mvad}}
\label{fig:EX}

\end{figure}
%order, example, group?
%Not sure, not enough space, write the other parts first.
Labels are not always clean, complete, and unequivocal. As illustrated in Figure~\ref{fig:EX} (top), given a training instance \inlineimg{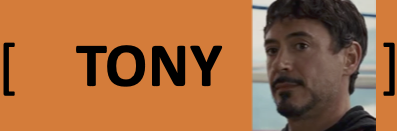}, it corresponds to a candidate label set [\inlinename{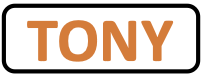},\inlinepe{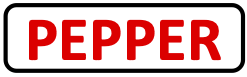}] where only one of them is correct. 
Learning from such ambiguous labels is known as Partial Label Learning (PLL)~\cite{cour2011PL}, which is a practical problem since it significantly reduces the human label effort compared to other one-to-one supervisions.

 \begin{figure*}[t]
 \centering
  \includegraphics[width=2.0\columnwidth]{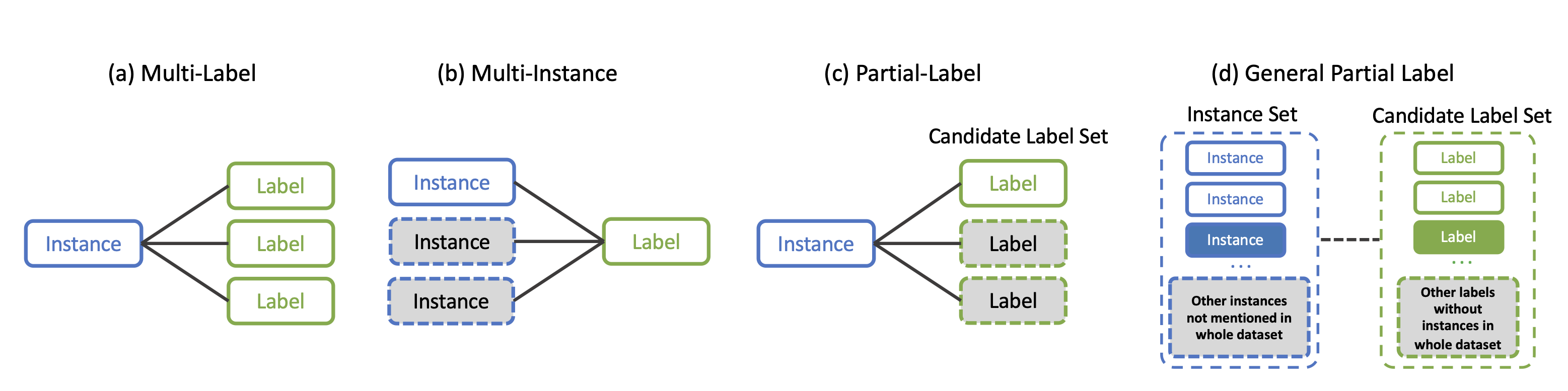}
  %\vspace{-5mm}
  \caption{From (a) to (d), on the evolution of relaxing the supervision but bringing in more label ambiguity challenges. \textbf{(a) Multi-Label}~\cite{huang2018fast}: each instance is labeled with more than one labels. \textbf{(b) Multi-Instance}~\cite{wu2015deep}: at least one instance in a group belongs to the label. %(a) and (b) can be combined as Multi-Instance Multi-Label ~\cite{zhou2012multi}. 
  \textbf{(c) Partial-Label}~\cite{feng2019partial}: a candidate label set partially labels an instance, only one of them is correct. Note that the label set may vary from instance to instance. So far, the supervision is on instance-level. \textbf{(d) General Partial Label Learning} (our focus): group-level supervision. Instances or labels (blue or green shaded) may link to another group; there are also \emph{null} instances and labels (grey shaded) with no links at all.}
  \label{fig:compare}
\end{figure*}

However, the assumption of PLL is still hardly feasible in large-scale scenarios: if we have millions of frames in videos or Web images%, or unaligned translation corpus
, the \emph{instance-level} label annotation of PLL will be prohibitively expensive.
%and hardly hold. % in real-world data. 
Figure~\ref{fig:EX} (bottom) shows several examples of the relaxation from instance-level to \emph{group-level}: a group of instances \inlineimg{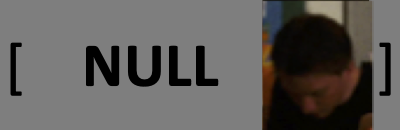} \inlineimg{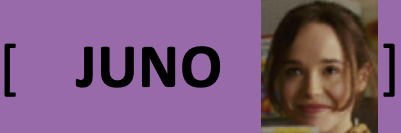} \inlineimg{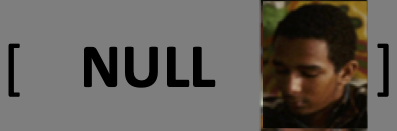} and the candidate label set [\inlinename{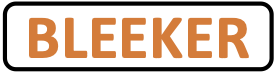},\inlinename{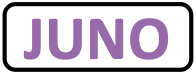}]. Compared to the tradition PLL, this is more complex and ambiguous: 1) \emph{within-group} annotations are dropped, 2) \emph{cross-group} links are allowed --- \inlineimg{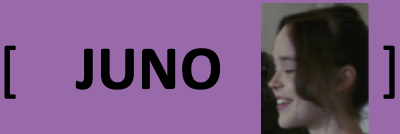} appears in the candidate label set of another group [\inlinename{1632juno_n}], and 3) there are some instances \inlineimg{1632null3} with $null$ label that is not in any label set. Such relaxed supervision is more appealing since it requires NO extra annotation on the instance-level. To this end, we propose a novel problem: General Partial Label Learning (GPLL), whose training annotation only comes from the inherent data pair (Figure~\ref{fig:EX} bottom), and thus is very challenging. Figure~\ref{fig:compare} illustrates some related problems with progressively relaxed supervisions.

A straightforward approach to GPLL is to consider some \emph{within}-/\emph{cross}-group heuristics such as : 1) Instances with similar features \inlineimg{1632juno} \inlineimg{1632juno2} across groups likely belong to the same label. 2) Similar instances \inlineimg{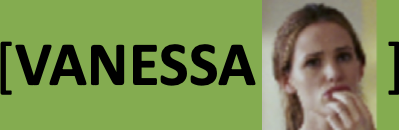} \inlineimg{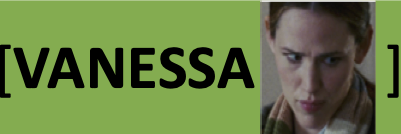} co-occur with the same label \inlinename{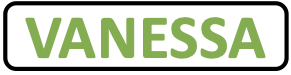} across groups implies that the label is likely assigned to those instances. 3) An instance cannot belong to multi-labels \inlineimg{1632van} $\rightarrow$  \inlinename{1632van_n}, \inlineimg{1632van} $\rightarrow$ \inlinename{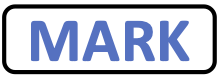}, and distinct instances \inlineimg{1632van} \inlineimg{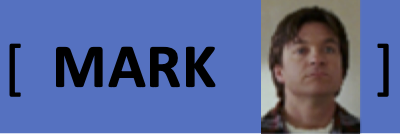} cannot be the same label \inlinename{1632mark_n} within a group. However, these heuristics are too weak to address the extreme ambiguity. In fact, as we will show in Ablation Study, modeling such heuristics to \emph{construct the initial links} only achieves 62.9\% accuracy.

We believe that the key to solve GPLL is how to exploit the aforementioned cross-group correlations unsupervisedly to construct initial links  and then \emph{refine} them with stronger group contextual representations. To this end, we propose a novel graph convolutional network, called \textit{\underline{D}ual \underline{B}ipartite \underline{G}raph \underline{A}uto\underline{e}ncoder} (DB-GAE). 
As its name implies, DB-GAE explicitly learns richer \emph{within-group} and \emph{cross-group} representations which serve as a reciprocal complement to each other. The within-group representation resolves the ambiguity in a group, and the cross-group one renders additional global group context for further disambiguation. In particular, we first represent the initial links as the proposed within-group and cross-group bipartite graphs, and then use GCN~\cite{berg2017graph} to encode and decode them to \emph{refine} the dual links to obtain the results, where the reconstruction loss is only referenced to the within-group graph input as this is the only supervision we have in GPLL. Therefore, it is worth noting that DB-GAE is self-supervised and transductive, which is appealing as it requires NO additional training data and an offline training stage.

%Summarize your experiments: datasets, performances, conclusions.
We compare the proposed DB-GAE to other baselines on both GPLL and PLL benchmarks. Our method outperforms the best baseline with absolute 0.158 F1-score and 24.7\% accuracy. We also analyze the model performances on the varying levels of label ambiguity. %and the amount of 
The contributions are summarized as follows:
\begin{itemize}
% \item We 
\item We introduce the new learning problem of GPLL, which generalizes the existing PLL formulation to more realistic, challenging, and ambiguous annotation scenarios.

\item We propose a novel graph neural networks called DB-GAE, which aims to disambiguate and predict instance-label links within and across groups.

\item We set up a new benchmark for the proposed GPLL task. Experiments demonstrate that DB-GAE significantly outperforms over strong baselines.

\end{itemize}

\begin{figure*}[t]
\centering
  \includegraphics[width=2.0\columnwidth]{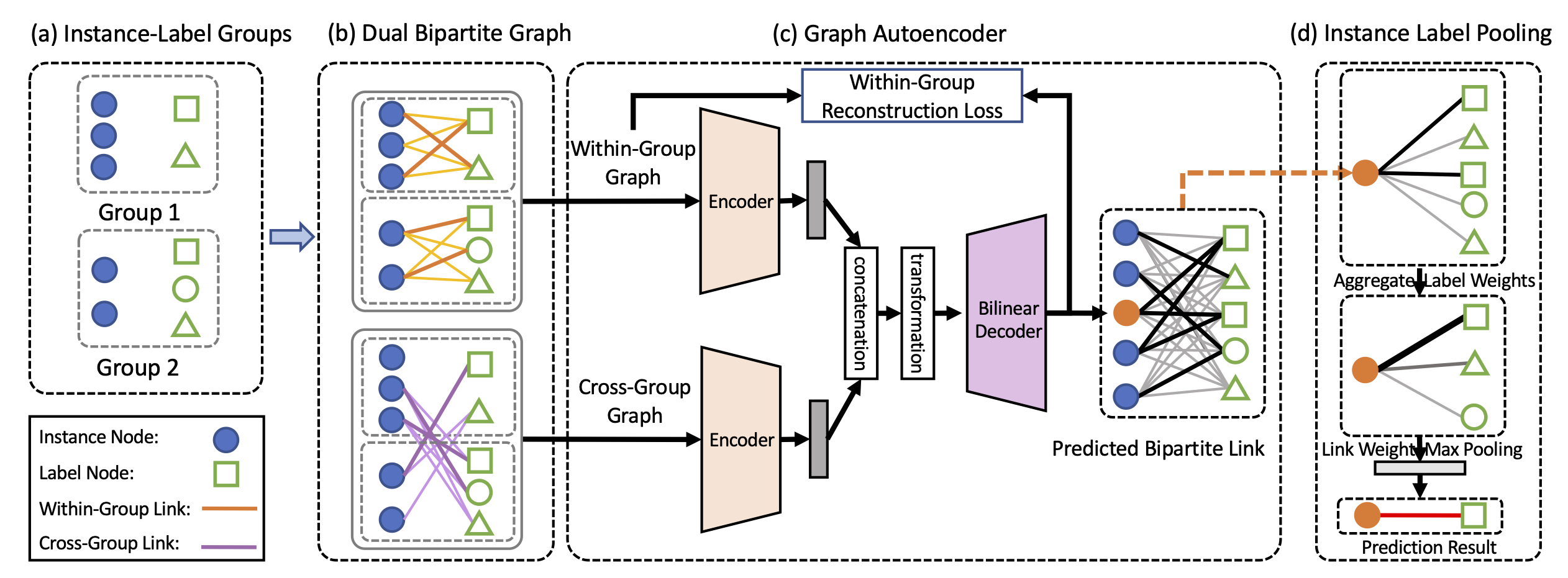}
  %\vspace{-8mm}
  \caption{The framework of the proposed method can be demonstrated as four parts: (a) Problem Formulation with Instance-Label Groups. (b) Dual Bipartite Graph. (c) Graph Autoencoder. (d) Instance Label Pooling.
  }
  \label{fig:framework}
\end{figure*}

%\vspace{-1mm}
\section{Related Work}

\noindent\textbf{Partial Label Learning (PLL)}~\cite{nguyen2008classification,cour2011PL,xie2018partial} also called superset label learning ~\cite{gong2017regularization} had been viewed as a weakly-supervised learning framework with implicit labeling information which assumes there is always exactly one ground-truth among the candidate label set. Therefore, one disambiguation strategy is building a certain parametric model and regarding ground-truth label as a latent variable. The model is iteratively refined by optimizing certain objectives, such as the maximum likelihood criterion~\cite{kupfer2019valuable,liu2014learnability}, or the maximum margin criterion~\cite{yu2016maximum}. 
Another strategy assumes equal importance for all kinds of candidate labels and predicts label scores by averaging their modeling outputs ~\cite{cour2011PL,tang2017confidence,wu2018towards,wang2019adaptive,xu2019partial}. Compared to the PLL problem, GPLL is much more challenging and needs to resolve group-level disambiguation, which is more general and practical for real-world scenarios.

%general ambiguity, without training data

\noindent\textbf{Graph Neural Networks (GNNs)} were introduced in~\cite{gori2005new,scarselli2008graph}, and mainly focus on supervised node classification or link prediction problem based on convolutional graph networks~\cite{defferrard2016convolutional,kipf2016semi,kipf2016variational,zhang2018link}. More recently, graph autoencoder networks~\cite{berg2017graph} were proposed to perform unsupervised link prediction, which we adopt for our label disambiguation problem. Unlike previous link prediction problems where the weights of observed links were given by the data. Our weights were initially estimated by clustering algorithms, which is the only information we have.

\iffalse
%\vspace{-3mm}
\begin{figure} [!th]
  \begin{center}
    \includegraphics[width=\columnwidth]{1632out.png}
  \end{center}
  %\vspace{-5mm}
  \caption{Example prediction output of the GPLL problem with the reference to Figure~\ref{fig:EX} (bottom). }
  \label{fig:out}
\end{figure}
%\vspace{-5mm}
\fi

\section{Problem Formulation}

In GPLL setting, the data is provided in the form of  $K$ groups $\mathcal{G}  =\{g_i\}^K_{i=1}$. Each group $g_i$ is a collection of instances and a candidate label set where $g_i = \{\mathcal{X}^{(i)},\mathcal{L}^{(i)}\}$. $\mathcal{X}^{(i)}$ is the set of $M^{(i)}$ instances, $\mathcal{X}^{(i)} = \{x^{(i)}_m\}^{M^{(i)}}_{m=1}$, and $x^{(i)}_m$ is the instance feature where $x^{(i)}_m \in \mathbb{R}^d$. %, \forall i = 1, . . ., N, m = 1, . . ., M_i$. 
The associated candidate label set $\mathcal{L}^{(i)}$ is the set of $N^{(i)}$ labels, $\mathcal{L}^{(i)} = \{l^{(i)}_n\}^{N^{(i)}}_{n=1}$,  %l^{(i)}_n \in \mathcal{Y}, %\forall i = 1, . . ., N, m = 1, . . ., N_i$, 
where $l^{(i)}_n \in \mathcal{Y}$.
The class set $\mathcal{Y}$ contains $(C+1)$ classes where $\mathcal{Y} = \{1, ,..., C, null\}$, since some instances might be from background classes that never appear in the dataset labels. As shown in the GPLL examples in Introduction, the correct label for an instance in $\mathcal{X}^{(i)}$ may exist in its candidate label set $\mathcal{L}^{(i)}$ or even in another candidate label set $\mathcal{L}^{(j)}$ of another group, where $i\neq j$. The instance in $\mathcal{X}^{(i)}$ will have a $null$ label if its correct label doesn't exist in any candidate label set $\mathcal{L}$ in $\mathcal{G}$.
In a nutshell, the input of this problem is a set of groups which consist of instances and labels $\mathcal{G}=\{\mathcal{X}^{(i)},\mathcal{L}^{(i)}\}^{K}_{i=1}$. The output is the predicted label $l \in \mathcal{Y}$ for each instance $x \in \mathcal{X}$. %Figure~\ref{fig:out} shows the output of this problem by the input data from Figure~\ref{fig:EX} (bottom) where the within-group labels, cross-group labels, and $null$ labels are correctly predicted. 
In addition, we assume that some instances and labels repetitively co-occur across groups for the model to learn the association pattern. Moreover, the problem is naturally in a \emph{self-supervised} and \emph{transductive} scenario, \ie, there is no train/test split, and the data $\mathcal{G}$ is all we have for labeling the instances.

%\vspace{-2mm}
\section{Approach}

As shown in Figure~\ref{fig:framework}, we describe our method for GPLL task and elaborate on each part with details: (b) In Dual Bipartite Graph, we take instances and labels as nodes and construct Within-Group or Cross-Group Link for two bipartite graphs with uncertain links. %learn to generate
(c) We propose a Graph Autoencoder to learn the embedding representations of the bipartite graphs and iteratively refine the bipartite link weights. 
(d) In the final stage, we propose an Instance Label Pooling to predict the correct instance-label link for each instance.

\subsection{Dual Bipartite Graph}  \label{sec:subgraphconstruction}
 Formally, we define our uncertain bipartite graph as $G = \{[\mathcal{X}, \mathcal{L}], M\} $, $G$ is a weighted graph with instance nodes $\mathcal{X}$ and label nodes $\mathcal{L}$. 
  $M$ denotes the uncertain links between instance $\mathcal{X}$ and label $\mathcal{L}$ with likelihood values. %probability value. Each probability value refers to the
  The likelihood of each link refers to whether a label is correct for an instance. We will construct the dual bipartite graph $M= [M^{within},M^{cross}]$ with complementary information. $M^{within}$ and $M^{cross}$ will be the edges of the Within-Group Graph and Cross-Group Graph.

\noindent\textbf{Within-Group Graph Construction.} 
As shown in Figure~\ref{fig:framework}(b), we consider the second and third heuristics described in the third paragraph of the Introduction to estimate the link likelihood within a group to construct the Within-Group Graph. 
The within-group link weight initialization contains three steps:
%\begin{itemize}
%\item
1) Given a instance $x_i \in \mathcal{X}$ and label $l_j \in \mathcal{L}$, we represent the within-group link $q_{ij}$ by concatenating instance and label features to form a tuple $[x_i;l_j]$. 2) We create all possible links within each of group between instances and labels and perform DBSCAN \cite{sander1998density} to 
cluster the links by their link features. We choose the cluster size $c_{ij}$ for each link to describe the co-occurrence frequency of instance-label pairs. 
The number is the times that $x_i$ co-occurs with the label $l_j$ in the entire dataset, which assigned as the likelihood of $l_j$ being the correct label of $x_i$.  %\bobby{need to be revised}
%\item
%Normalize by contradictory link weights.
3) We will refine the likelihood by considering the contradictory relation of links within a group.%, as shown in 
We define the contradictory link for each link $q_{ij}$: the links with only one shared node (instance $x_i$ or label $l_{j}$) in the same group. We will refine the likelihood by dividing the total likelihood of the link $q_{ij}$ and its contradictory links. The within-group link weight is defined as follow: 
\begin{equation}
w_{ij} = \frac{c_{ij}}{(\sum_{u\in N_i}c_{iu}+\sum_{v\in N_j}c_{vj})-c_{ij}}
\end{equation}
where $N_i$ and $N_j$ are the neighbor nodes set for node $i$ and $j$. 
We acquire all the within-group link weights by calculating all the weights between the instances and labels within the same group and denote the weighted adjacency matrix as $M^{within}\in [0,1]^{U \times V}$ where $U$ is the number of instances in $\mathcal{X}$ and $V$ is the number of labels in $\mathcal{L}$.

\noindent\textbf{Cross-Group Graph Construction.}
Given the within-group weights $M^{inner}$ and the first heuristic mentioned in the third paragraph of Introduction, we can initialize the cross-group link weight and construct Cross-Group Graph shown in Figure~\ref{fig:framework}(b).
Given an instance, 
we measure the l2-distance for instances
and select similar instances by a certain threshold $d$ and define those nodes as homogeneous neighbor node. %\bobby{and define 
In addition, the homogeneous neighbors of each instance has their candidate labels, and we link the instance to these cross-group candidate labels as a cross-group link. The likelihood values of cross-group links were initialized in the previous step. 
We use such within-group link weight to be the likelihood between the instance and the label of its homogeneous node.

\subsection{Graph Autoencoder}  

To predict the unknown likelihood of instance-label pairs for uncertain graph, we design a novel graph autoencoder architecture called DB-GAE. The model has the ability to 1) Encode the graph $G$ with heterogeneous within/cross-group links to a low-dimensional embedding space. 2) Dynamically update the instance-label relation while learning a new representation of the instance and label nodes. 3) Predict the link weights between instances and labels by reconstructing the observed links we initialize.

\noindent\textbf{Graph Convolution Encoder.}
Given the node features $[\mathcal{X},\mathcal{L}]$ and the link weights $[M^{within}, M^{cross}]$ initialized in the previous step, we aim to encode such information in node representation for further prediction.
The graph convolution model incorporates the neighbor information by propagating the message to form a new representation of a node. We utilized this characteristic to use the link information during propagation to obtain a more representative embedding. 
For expressing the propagation of within-group links, a single hidden layer GCN is given by
\begin{equation}
\boldsymbol{H}^i=f(\boldsymbol{H}^{i-1},M^{within})
\end{equation}
where $\boldsymbol{H}^0 = [\mathcal{X},\mathcal{L}]$ and $f$ is a propagation rule. Each layer $\boldsymbol{H}^i$ corresponds to the instance and label feature matrix $[\mathcal{X}, \mathcal{L}]$ and where each row is a feature representation of a node. 
This operation is similar to a filtering operation in the CNN ~\cite{lecun1995convolutional}, and the features become increasingly more abstract at each consecutive layer. We aggregate the feature representation of each node by its associated neighbors. Moreover, the neighbors were weighted by the within-group weight $w_{ij}$ and transformed by applying the weights $W$ before propagation. 
 To avoid the interference between within-group and cross-group link propagation, we have distinct propagation rules of dual bipartite GCNs for within-group links and cross-group links separately.
 The propagation rule for within-group and cross-group can be denoted as: 
\begin{equation}
\mu_{j \rightarrow i} = w_{ij} W l_j  \quad
\mu_{j^{\prime} \rightarrow i} = w_{ij^{\prime}} W l_{j^{\prime}} 
\end{equation}
where the $w_{ij}$ and $w_{ij^{\prime}}$ are link weights computed from the previous section and $j^\prime$ is a cross-group label. $W$ is a learnable parameter.
This operation is similar to the spectral rule propagation \cite{kipf2016semi} where the propagation is normalized based on the degree of both $i$ and $j$. Instead, our propagation is normalized by the link weight of both $i$ and $j$. 
We aggregate incoming messages for each of instance from label nodes by accumulating all neighbors $N_{i}$ to represent the node, denoted as:
\begin{equation}
h_i^{within} = \sigma (\sum_{j \in N_i} \mu_{j \rightarrow i})  \quad
h_i^{cross} = \sigma (\sum_{j^{\prime} \in N_i} \mu_{j^{\prime} \rightarrow i})  
\end{equation}
$h_i^{within}$ is the hidden vector that represents the instance node $i$ by within-group and $h_i^{cross}$ is the hidden vector that represents the instance node $i$ by cross-group. 
To arrive at the final embedding of instance node $i$ and label node $j$, we apply the concatenate operation over the hidden vector updated from the within-group and cross-group.
The model has a non-linear transformation to transform the concatenated representations for each node by dual path GCN to a unified embedding representation. After concatenation, the feature will feed into a dense layer to obtain the final representation, denoted as:
\begin{align}
    u_i =  \sigma (W_u[h_i^{within};h_i^{cross};f_i])  \\
   v_j = \sigma (W_v[h_j^{within};h_j^{cross};n_j])
\end{align}
%$f_i$ is the input instance and label feature. 
$\sigma (\cdot)$ denotes an ReLU activation function. $W_u$ and $W_v$ are learnable parameters.
We use the transformation functions of $f_i = \sigma(W_f x_i+b)$ and $n_j = \sigma(W_nl_j+b)$ in our paper. The output of the encoder will be the updated representations $[\boldsymbol{U,V}]$ for the instances and labels. %\bobby{add the sentence}.

\noindent\textbf{Graph Attention for Within/Cross-Group Propagation.}
%\bobby{mark}
The graph convolution is based on the probability value of in the dual bipartite graph, which is fixed during the graph propagation process. Moreover, we want to continuously update the representations of nodes to predict the link weights by learning from links with uncertainty. Hence, dynamically adjust the propagation weight between instances and labels by considering the features itself is essential. 
To this end, we can 
employ some form of attention mechanism \cite{velivckovic2017graph} which actively learn how to propagate the information to optimize our result. %Moreover, the individual contribution of each message is learned and determined by the model. 
To perform the attention on nodes, attention coefficients can be calculated by: 

\begin{equation}
e_{ij} = a(W_{a}x_i,W_{a}l_j)  
\end{equation}
It indicates the importance of label node $l_j$'s features to instance node $x_i$, and $W_a$ is its learnable weight matrix, and $a$ is a feed-forward network. We inject the graph structure into the mechanism by performing masked attention which means we compute $\alpha_{ij}$ for nodes $j \in  N_i$, where $N_i$ is the neighbor nodes of node $i$ in the graph.
To make coefficients easily comparable across different nodes, we normalize them across all choices of $j$ using the softmax function:
\begin{equation}
\alpha_{ij} = \frac{exp(e_{ij})}{\sum_{k \in N_i} exp(e_{ik})}  
\end{equation}

\noindent We learn two kinds of link information by graph attention, including the within-group link weight and cross-group link weight. 
Therefore, the propagation rule in Equation 3 can be extended to:
\begin{equation}
\mu_{j \rightarrow i} = \alpha_{ij} w_{ij} W l_j  \quad
\mu_{j^{\prime} \rightarrow i} = \alpha_{ij^{\prime}} w_{ij^\prime} W l_{j^{\prime}} 
\end{equation}
where $\alpha$ is the context vector which represents the normalized contribution of label $j$ to instance $i$. 
After aggregating the information though the neighbors by summation and apply
\iffalse
, the Equation 3 becames: 
\begin{equation}
h^{ within}_i = \sigma (\sum_{j \in N_i} \mu_{j \rightarrow i} )  \quad 
h^{ cross}_i = \sigma (\sum_{j^{\prime} \in N_i} \mu_{j^\prime \rightarrow i} )
\end{equation}
$h^{\prime within}_i$ and $h^{\prime outer}_j$ are the updated feature after multiplying the attention coefficient.

To stabilize the learning process, we employ multi-head attention \cite{zitnik2017predicting} to be beneficial. Specifically, $K$ independent attention mechanisms execute the transformation of Equation 9, and then their features are concatenated. 
Later, we apply 
\fi
averaging on the $K$ transformation attention by multi-head attention \cite{zitnik2017predicting}, the Equation 4 becomes: %\bobby{I know this sentence is for showing the extended equations， so maybe we can just keep the equations of the last step}:
\begin{align}
h^{ within}_i = \sigma(\frac{1}{K}\sum_{k=1}^K \ \sum_{j \in N_i} \mu^k_{j \rightarrow i}) \\ 
h^{ cross}_i = \sigma(\frac{1}{K}\sum_{k=1}^K \ \sum_{j^\prime \in N_i} \mu^k_{j^\prime \rightarrow i})
\end{align}

\noindent\textbf{Bi-linear Decoder for Link Resolution.}
To predict the link values between instances and labels, we decode the updated embedding that contains the within-group, cross group, and feature information.
In addition, we can use Bi-linear decoder model~\cite{kiros2014unifying} to reconstruct links of the bipartite graph by considering the node feature similarity.
The reconstruction model is $M = \sigma ( \boldsymbol{U}^{T}\boldsymbol{V})$, and likelihood between instance $i$ and label $j$ is $\hat{M}_{ij}$. The learning objective is to reconstruct the weights of the observed links (estimated from the within-group initialization) and predict the weights of unobserved links (cross-group link).  The score function of the decoder is:
\begin{equation}
p(\hat{M}_{ij}=r) = \frac{e^{u^T_i Q_r v_j}}{\Sigma_{s\in R}e^{u^T_i Q_s v_j} },
\end{equation}
where $Q_r$ is the trainable parameter matrix of shape $E \times E$, and $E$ is the dimension of hidden representations. $r$ is weighting scale from 0 to 1 which represents the likelihood of the link.
The predicted rating is computed as:
{\footnotesize{
\begin{equation}
\hat{M}_{ij} = g(u_i,v_j) = \mathbb{E}_{p(\hat{M}_{ij}=r)}[r] = \sum_{r \in R} r p(\hat{M}_{ij}=r)
\end{equation}}
}

\noindent \textbf{Within Group Reconstruction Loss.} %Objective Function.} 
 To optimize the proposed graph inference networks, we follow the loss function defined in \cite{berg2017graph} to minimize the reconstruction loss by negative log likelihood of the predicted likelihood:
 \begin{equation}
 \mathcal{L} = - \sum_{i,j;\Omega_{ij}=1}\sum_{r=1}^{R} I[r = M^{within}_{ij}] \log p(\hat{M}_{ij} = r),
 \end{equation}
 %equation [XX]
where the matrix $ \Omega \in {0, 1}$ is a mask for unobserved links in the within-group matrix $M^{within}$. We optimize over the observed links to predict the likelihood of the matrix $\hat{M}$ which contains observed links and unobserved links. %\bobby{need rephrase}.

\subsection{Link Prediction by Instance Label Pooling}
To infer the label of each instance, we use the predicted link weight $\hat{M}_{ij}$ generated by DB-GAE. 
As shown in Figure\ref{fig:framework}(d), %the DP-GIN can work as a link refinement for the within-group links and also link weight prediction for cross-group links \bobby{mark}. G
given an instance, we aggregate all the weights of the links it connects by different classes. % (every label belongs to a class). 
For within-group link weight, we directly use the predicted link weight. For cross-group link weight, we multiply the predicted link weight with the cosine similarity of the instance $i$'s feature and its homogeneous neighbor's feature. That is because the link weight should be lower if the feature similarity is low. The weight of a class $o$ being the label of instance $i$ is calculated by:
 \begin{align}
W_o^i =  \sum_{j \in o;\Omega_{ij}=1}\sigma(\hat{M}_{ij}) + \sum_{j \in o;\Omega^{\prime}_{ij^{\prime}}=1}\sigma(\hat{M}_{ij^{\prime}}\frac{f_i \cdot f_{i^\prime} }{||f_i||||f_{i^\prime}|| } )
 \end{align}
 where the $\sigma$ represents a ReLU over a sigmoid function. $ \Omega^\prime \in {0, 1}$ is a mask for unobserved links in the cross group matrix $M^{cross}$.
 We aggregate the link weight by the same class and pool the class with the maximum weighted score $p_i = argmax_{o \in \mathcal{Y}}(W^i_o)$ as the predicted label of instance $x_i$. 
 If the score is equal to 0, it means there is no prediction, we predict it as a $null$. %We 

%\vspace{-3mm}
\subsection{General Partial Label Learning Datasets} \label{sec:dataset}
% delete some words

We evaluate the performance of our model for the automatic face naming problem on two real-world datasets: MPII-MD~\cite{RohrbachCVPR2017} and M-VAD \cite{pini2019mvad}. MPII-MD dataset in GPLL setting, which contains only group-level supervision with cross group labels and $null$ labels. The M-VAD dataset is constructed for PLL setting, which has less ambiguity but larger-scale data.

\noindent\textbf{MPII-MD: MPII Movie Description Dataset.}
The MPII-MD dataset consists of face images and ambiguous labels automatically extracted using the screenplays provided in 13 movies with 806 different faces and 181 possible names from 558 image-caption pairs. We select the frames from the data with detectable faces and corresponding captions. The percentage of the faces with a $null$ label in the dataset are 21\%.

\noindent\textbf{M-VAD: Montreal Video Annotation Dataset.}
In M-VAD Names dataset contains fully annotated faces in images with names in the captions from 55 movies. It consists of 222,58 detected faces and 591 possible names from 17,533 image-caption pairs.

\begin{figure}[!th]
  \includegraphics[width=\columnwidth]{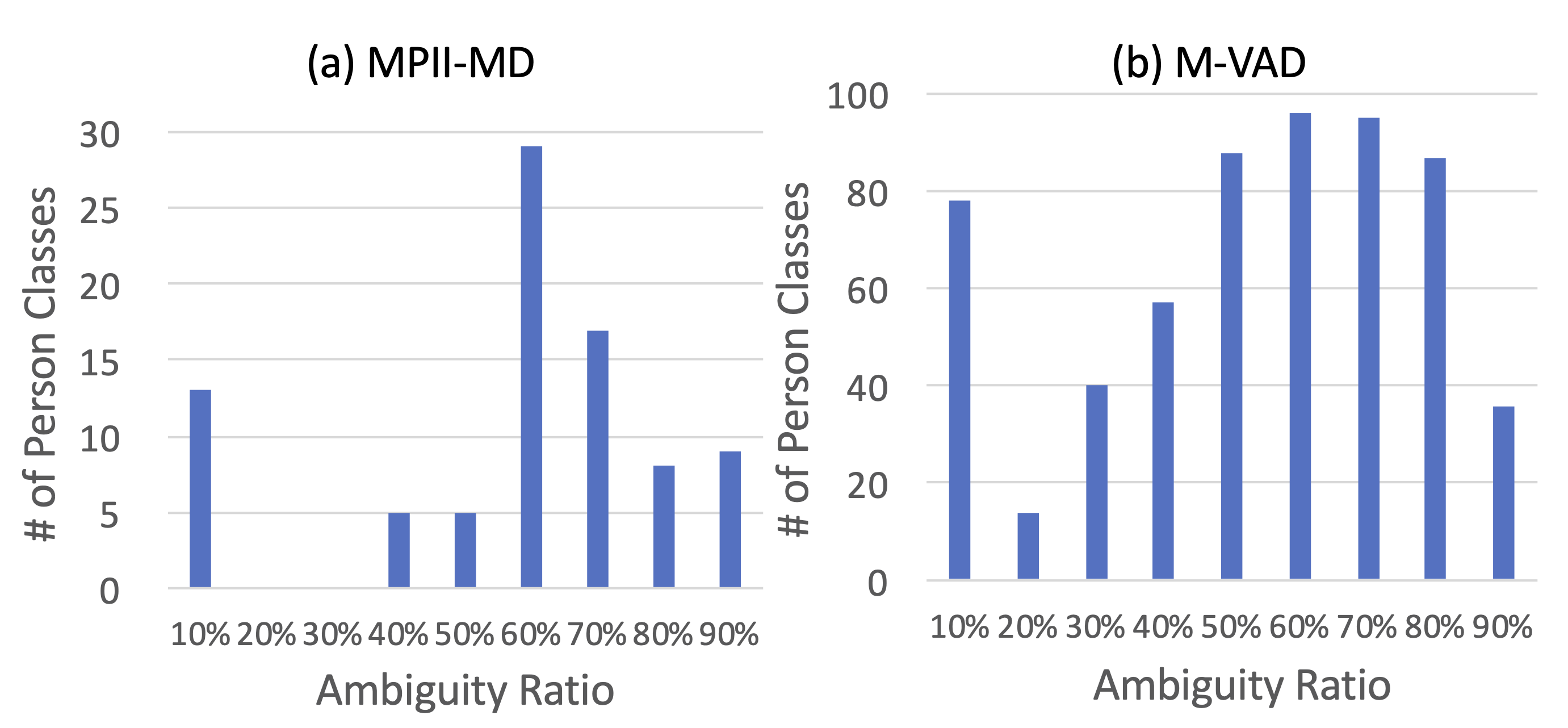}
  %\vspace{-7mm}
  \caption{Data Distribution over Ambiguity Ratio}
  \label{fig:dataratio}
\end{figure}

\noindent\textbf{Data Distribution over Ambiguity Ratio.}
To explore the data difficulty of the datasets, we define \textit{ambiguity ratio} and show the histogram over different levels of ambiguity, as shown in Figure~\ref{fig:dataratio}. The metric refers to a fraction of all possible instance-label links that are incorrect, the ambiguity ratio of a label with class $c \in \mathcal{Y}$ is defined by:
% \bobby{what is the sentence meaning}:
%\vspace{-3mm}
\begin{equation}
\mathcal{R}_o =  1- \frac{ \sum_{i=1}^{K} |s^{(i)}_t|}{\sum_{i=1}^{K} |s^{(i)}_t| + |s^{(i)}_f|}
\end{equation}
where $i$ is the group index. $s_t^{(i)}$ is the set of correct instance-label links in the group $g^{(i)}$ with class $o$. $s_f^{(i)}$ is the set of wrong links in group $g^{(i)}$ which connect to instance node or label node with class $o$. 

\begin{table*}
  \caption{Performance Comparison on MPII-MD and M-VAD}
  \label{sample-table}
  \centering
  %\caption{Performance Comparison with Different Baseline Methods.}
  \resizebox{\columnwidth}{!}{
  \begin{tabular}{lccc}
    \toprule
    \hline
    \multicolumn{1}{c}{} &  
    \multicolumn{2}{c}{MPII-MD} &                   
    \multicolumn{1}{c}{M-VAD}     \\               
    \hline
    Method     & F1-score     & Accuracy  & Accuracy\\
    \midrule
    Pair Clustering  & 0.539  & 45.0 \% & 78.2 \%   \\ 
    Cluster Voting & 0.558  & 46.5 \% & 78.9 \%   \\
    %Label Propagation & 0.679  & 0.598  & 0.816   \\
    \hline
    SURE  & 0.605  & 48.7 \% &  85.7 \%   \\
    IPAL  & 0.608  & 48.3 \%  & 86.1 \%  \\
    PL-LEAF & 0.610  & 48.6 \% & 86.3 \%   \\
    PL-AGGD  & 0.598  & 47.6 \% & 86.5 \%  \\
    \hline
    Our method & \textbf{0.768} & \textbf{76.5 \%} & \textbf{90.3 \%}   \\
    \bottomrule
  \end{tabular}
  }
 \end{table*}
 % \end{minipage}

%\vspace{-3mm}
\subsection{Baselines}

\noindent\textbf{Cluster Voting~\cite{sander1998density}:} 
For each instance, the method selects
candidate labels from the same cluster.  The correct label is determined by majority voting over all the candidates. To cluster faces by visual features for face naming datasets, we apply DBSCAN (with $\epsilon = 1, n=2$).

\noindent\textbf{Pair Clustering~\cite{sander1998density}:}
As in the Within-Group Graph Construction, we perform pair clustering to estimate the likelihood of a link. Given an instance, we find its link with the largest cluster size and pick its label as the prediction, We also perform DBSCAN (with $\epsilon = 1, n=2$) for pair clustering.

\noindent\textbf{IPAL \cite{zhang2015solving}}: IPAL is an instance-based PLL model and disambiguates candidate labels by an iterative label propagation procedure.

%  an iterative label propagation procedure. During the testing phase, the unseen instance is classified based on minimum error reconstruction from its nearest neighbors

\noindent\textbf{PL-LEAF \cite{zhang2016partial}}: PL-LEAF is a feature-aware approach which learns the manifold structure of feature space and performs regularized multi-output regression over the generated labeling confidences.

\noindent\textbf{SURE \cite{feng2019partial}}: SURE proposes a unified formulation with the maximum infinity norm regularization to train the desired model and perform pseudo-labeling jointly.

\noindent\textbf{PL-AGGD \cite{wang2019adaptive}}: PL-AGGD proposes a unified framework which jointly optimizes the ground-truth labeling confidences, similarity graph, and model parameters to achieve generalization performance.

%\vspace{-3mm}
\subsection{Experimental Setup}
For all runs, we use pre-trained FaceNet \cite{schroff2015facenet} to extract the visual embedding for detected faces with 512 dimensions from the images and apply a threshold $d=1$ suggested in the paper \cite{schroff2015facenet} for $l_2$ distance to determine two faces are the same person. We encode names using one-hot vectors. %We use this threshold on DBSCAN face clustering, pair clustering, and determine the homogeneous neighbors. 
In DB-GAE, we use the Adam optimizer \cite{kingma2014adam} with a learning rate of 0.001. The layer sizes of graph convolution (with ReLU) is 1000 and 100 for the dense layer. We run 1000 epochs on both datasets with the runtime of 10min on MPII-MD dataset and 5hr 10min on the M-VAD dataset on CPU.
 Since M-VAD dataset has sufficient training data, we perform 10-fold cross-validation on the partial label learning method with 9:1 train/test split. For our method, we only use the same testing data (10\% data) because our method does not require additional training data. More details of the model architecture, parameter list, and analysis of the time complexity of the model will be included in the arXiv version.

%\vspace{-3mm}
\subsection{Methods Performance Comparison}
% \bobby{already rewrote the part}

From Table \ref{sample-table}, our proposed method outperforms the baselines in both datasets with a significant margin by 0.158 absolute improvements of F1-score for GPLL dataset and 3.8\% absolute improvement of accuracy in PLL dataset. %\bobby{use XX\% for performance improvements here}. 
In the experimental results on MPII-MD dataset, the best baseline reach about 0.61 on F1-score and 49\% on accuracy, and proposed model can achieve the best performance with over 0.76 on F1-score and 73\% on accuracy. The significant improvements show that our model is powerful enough to deal with generalized ambiguity, and the baseline methods will fail to disambiguate distractors. %The result difference between the Label Propagation and our method also shows that refining the link weight by the DB-GAE is essential to determine the name of each face. 
As shown in the results on M-VAD dataset, PLL methods performed much better when using sufficient training data for PLL setting. Our model is able to achieve the best performance among the methods with only one-tenth of data in the transductive setting. This result shows its ability to resolve the extreme ambiguity caused by data sparsity.

\begin{figure}[!th]
 \centering
  \includegraphics[ width=1\columnwidth]{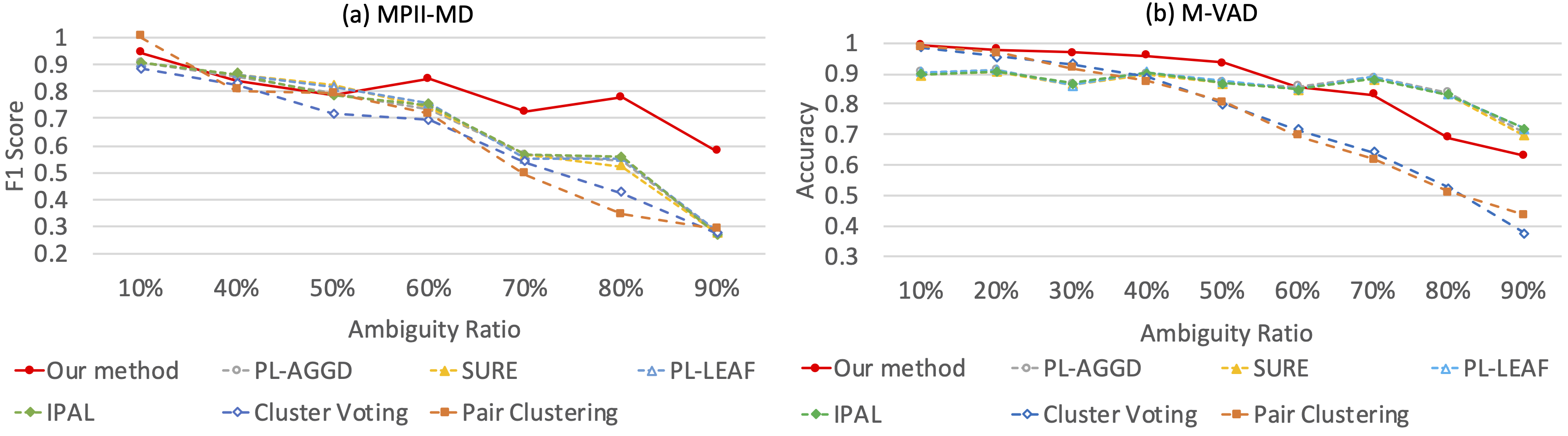}
  %\vspace{-7mm}
  \caption{F1-score and accuracy curves versus ambiguity ratios on MPII-MD and M-VAD. }
  \label{fig:ratio}
\end{figure}

\begin{figure}[!th]
\centering
 
  \includegraphics[width=1\columnwidth]{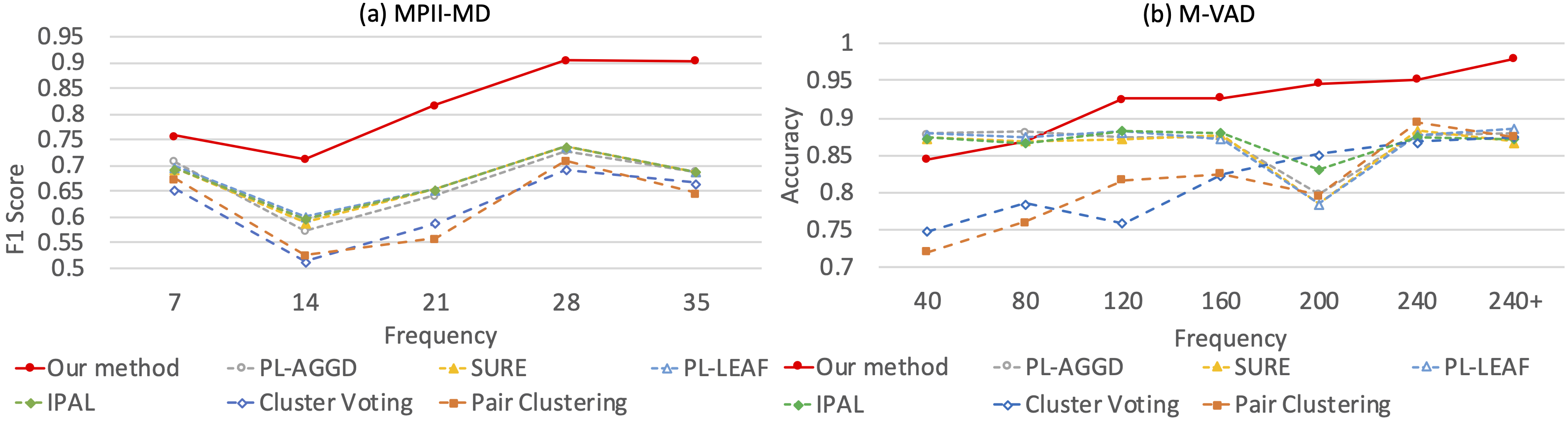}
  \caption{F1-score and accuracy curves versus ground-truth frequency on MPII-MD and M-VAD.}
  \label{fig:freq}
\end{figure}

%\vspace{-3mm}
\subsection{Condition Controlled Experiments}
% \bobby{already rewrote the part}

In addition to overall performance results, we show the comparison with baselines for different levels of data difficulty. %From the definition, we know that the higher the ambiguity ratio is, the data became more difficult. 

\noindent\textbf{The effect of ambiguity ratio on performance.}
In the Figure~\ref{fig:ratio}(a), the baselines can reach comparable performances with the proposed model when the ambiguity is not severe (ambiguity ratio $<$ 0.4). Their performances will drop a lot when they meet high levels of data ambiguity (ambiguity ratio $>$ 0.4). In less ambiguity dataset (PLL setting) Figure~\ref{fig:ratio}(b), we can see that the proposed method is comparable with the state-of-the-art PPL method without addtional training data.

\noindent\textbf{The effect of ground-truth frequency on performance.}
The \textit{ground-truth frequency} is the number of face-name pairs with the same class co-occur in the same group throughout the dataset.
% We also compare our method on different frequencies. We know that if the person has more frequent correct face-name pairs, the task will be easier. 
In Figure~\ref{fig:freq} (a), we can see that our model performs better than other methods in general. Almost all of the methods exhibit a slight drop in the ground-truth frequency of $14$, because the average ambiguity ratio at frequency interval is higher than the middle ambiguity ratio of frequency $7$. In Figure~\ref{fig:freq}(b), the approach performs better than other methods. The accuracy of the model will continually increase even beyond 95\% if the correct face-name co-occur frequently enough.

%\textbf{Win:Tie:Loss}
%\subsection{Further Analysis}

%\begin{comment}
\begin{table}
  \caption{Ablation Study of Proposed Method DB-GAE}
  \label{sample-table2}
  \centering
  \resizebox{\columnwidth}{!}{%
  \begin{tabular}{lccc}
    \toprule
    \hline
    \multicolumn{1}{c}{} & 
    \multicolumn{2}{c}{MPII-MD} &                   
    \multicolumn{1}{c}{M-VAD}    \\               
    \hline
    Method     & F1-score     & Accuracy      & Accuracy \\
    \midrule
    DB-GAE &\textbf{0.768} & \textbf{76.5\%} & \textbf{90.3\%}  \\
    w/o Graph Autoencoder & 0.614  & 62.9\%    & 81.6\%  \\
    w/o Dual Bipartite Graph & 0.724  & 68.4 \% & 88.3 \% \\
    w/o Cross-Group Link  & 0.730  & 73.9 \%    & 89.2 \%  \\ 
    w/o Graph Attention  & 0.743  & 76.4 \%      & 87.3 \%  \\
    \bottomrule
  \end{tabular}
  }%
\end{table}
%\vspace{-3mm}
% \iffalse
\subsection{Ablation Study}
% \bobby{already rewrote the part}

To reveal the contribution of each component, we test the performance of DB-GAE by removing different parts: Graph Autoencoder, Dual Bipartite Graph Architecture (apply GAE and set all weights for instance label links to be averaged \cite{hullermeier2006learning,zhang2015solving}), Cross-Group Link, and Graph Attention.
The comparison results of ablation study are shown in the Table~\ref{sample-table2}, we can see that the Graph Autoencoder contributes the most of performance improvement. The GAE (w/o Dual Bipartite Graph Architecture) encounters an accuracy drop in the MPII-MD dataset because it is unable to deal with null distractors since the weights fed into the GAE are all links with a high likelihood. 
%In the M-VAD dataset, there are no null distractors in the dataset. As a result, the accuracy drop is not that significant. Moreover, since the M-VAD names dataset has enough data to learn the pattern, the model doesn't benefit much on the weight initialization method.
The consideration of Cross-Group Link will help the model to deal with the group-level ambiguity. The F1-score of DB-GAE (w/o Cross-Group Link) drops obviously on MPII-MD. The improvement of DB-GAE over DB-GAE (w/o Graph Attention) verifies our hypothesis that graph attention can help to capture a better representation of the graph structure.
% \fi

\begin{figure}[!th]
  \includegraphics[width=1\columnwidth]{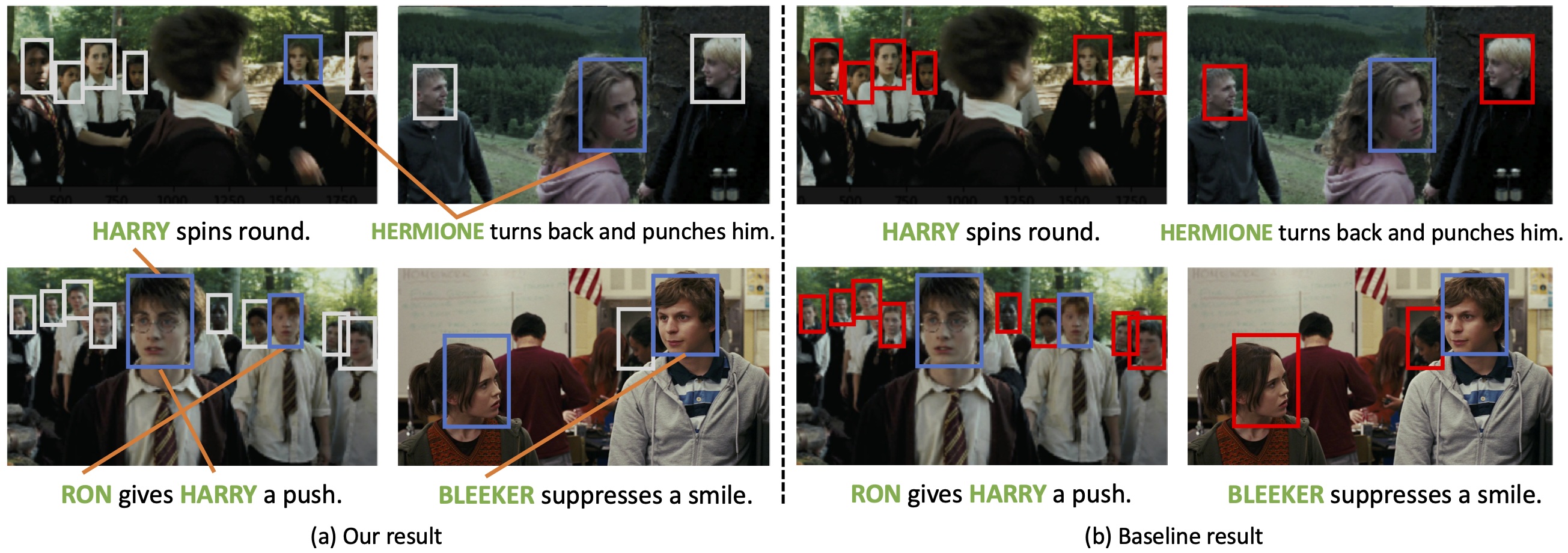}
  \centering
  %\vspace{-7mm}
  \caption{Qualitative examples from MPII-MD dataset. The blue/grey box represents the correct prediction of a name/null label. Red box represents the wrong prediction. The orange line visualize the predicted link between face and name.}
  \label{fig:qualitative}
\end{figure}

\begin{figure}[!th]

  \includegraphics[width=1\columnwidth]{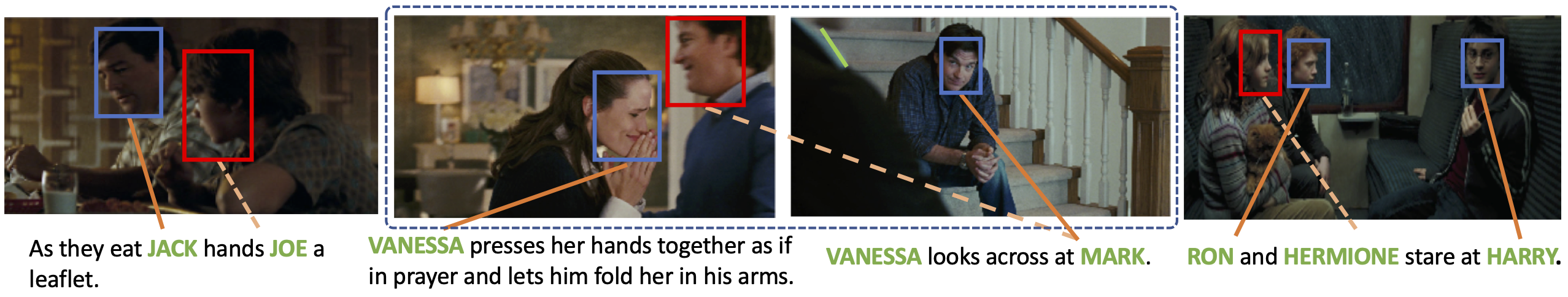}
  \centering
  %\vspace{1mm}
    %\vspace{-7mm}
  \caption{Failure cases from MPII-MD dataset. The dashed line represents the ground-truth link between face and name which the model misses.}
  
  \label{fig:fail}
\end{figure}

\subsection{Qualitative Results}
As shown in Figure~\ref{fig:qualitative}, compared to the best baseline: PL-LEAF in our MPII-MD experiment, our model can correctly predict the $null$ labels and cross-group labels even when the number of distractors is large within a group. Also, the performance for within-group label prediction is also better since the model incorporates cross-group knowledge to resolve the within-group ambiguity.

\subsection{Failure Cases}
From Figure~\ref{fig:fail}, we can conclude that
1) The visual recognition rate may affect the within-group linking performance. In the first example, \inlineimg{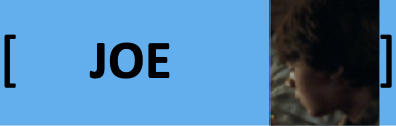} didn't link to \inlinename{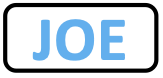}, which was limited by the feature representation. 2) Also, it will affect the cross-group linking due to the failure of finding similar faces. \inlineimg{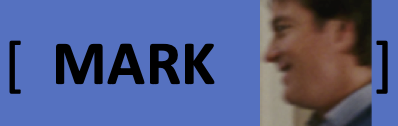} should link to the name \inlinename{1632mark} to find its correct label \inlinename{1632mark_n} but the model can't find \inlineimg{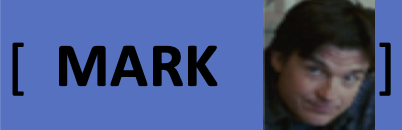} as similar faces. 3) The current model is based on correlation and thus lacks reasoning ability, for example, we humans may rule out other faces and predict the correct link, but our method fails. For example, when predicting \inlineimg{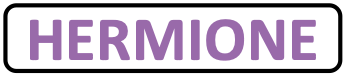}, since we know \inlinename{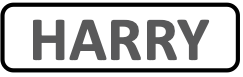} and \inlinename{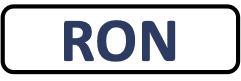} were linked, we can infer \inlineimg{1632her} is more likely to be \inlinename{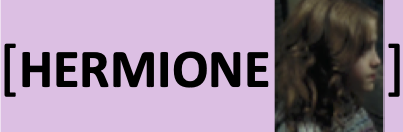}.

\section{Conclusions}
% \bobby{already rewrote the part}

% three point and solved
We introduced the General Partial Label Learning (GPLL) problem, which is more realistic and general than the traditional PLL. The proposed approach DB-GAE was designed to tackle the challenges of GPLL by disambiguating the within-/cross-group instance-label links with richer contextual graph representations. We contributed two GPLL benchmarks on automatic face naming tasks. We found that DB-GAE outperformed the best baseline with absolute 0.159 F1-score and 24.8\% accuracy.  Further analysis shows the robustness of DB-GAE in generalized ambiguity scenarios and the effect of various ambiguity levels. Moving forward, we are going to frame more tasks into GPLL such as cross-domain co-reference resolution in NLP, and push the envelope of DB-GAE in other fields.

\section{Acknowledgment}
This work was supported by the U.S. DARPA AIDA Program No. FA8750-18-2-0014. The views and conclusions contained in this document are those of the authors and should not be interpreted as representing the official policies, either expressed or implied, of the U.S. Government. The U.S. Government is authorized to reproduce and distribute reprints for Government purposes notwithstanding any copyright notation here on.


% Use unnumbered third level headings for the acknowledgments. All acknowledgments
% go at the end of the paper. Do not include acknowledgments in the anonymized
% submission, only in the final paper.

%\end{comment}

\bibliographystyle{aaai}
\bibliography{AAAI-BrianC.1632}

\end{document}